\documentclass[pdflatex,sn-nature]{sn-jnl}

\usepackage[T1]{fontenc}
\usepackage{lmodern}
\usepackage{graphicx}
\usepackage{amsmath}
\usepackage{xcolor}
\usepackage{textcomp}
\usepackage{array}
\usepackage{tabularx}
\usepackage{colortbl}

\makeatletter
\renewcommand{\email}[1]{\global\advance\emailcnt by 1\relax
\if@corauemail\g@addto@macro\corrauthemail{\setcounter{footnote}{0}\textcolor{blue}{#1}}
\else\g@addto@macro\authemail{\setcounter{footnote}{0}\textcolor{blue}{#1}}\fi}
\makeatother

\begin{document}

\title[The BRIDGE dataset]{The BRIDGE dataset: paired brightfield whole-slide images of prostate core-needle biopsies in paraffin, after deparaffinization and H\&E staining}

\author*[1]{\fnm{Julia A.} \sur{Mielcarz}}\email{julia.mielcarz.jm@gmail.com}

\author[2,3]{\fnm{Justyna} \sur{Durślewicz}}

\author[2]{\fnm{Marek} \sur{Zdrenka}}

\author[1]{\fnm{Ahmad} \sur{Abbadi}}

\author[1]{\fnm{Martin} \sur{Eklund}}

\author[2,3]{\fnm{Mateusz} \sur{Matusiak}}

\author[2,4]{\fnm{Łukasz} \sur{Szylberg}}

\author[1]{\fnm{Andrea} \sur{Camilloni}}

\affil*[1]{\orgdiv{Department of Medical Epidemiology and Biostatistics}, \orgname{Karolinska Institutet}, \orgaddress{\city{Stockholm}, \country{Sweden}}}

\affil[2]{\orgdiv{Department of Tumor Pathology and Pathomorphology}, \orgname{Prof. Franciszek Łukaszczyk Oncology Centre}, \orgaddress{\city{Bydgoszcz}, \country{Poland}}}

\affil[3]{\orgdiv{Center for Translational Pathology, Faculty of Medicine}, \orgname{Bydgoszcz University of Science and Technology}, \orgaddress{\city{Bydgoszcz}, \country{Poland}}}

\affil[4]{\orgdiv{Department of Obstetrics, Gynaecology and Oncology}, \orgname{Ludwik Rydygier Collegium Medicum in Bydgoszcz, Nicolaus Copernicus University in Toruń}, \orgaddress{\city{Bydgoszcz}, \country{Poland}}}

\abstract{Computational approaches are increasingly being explored to streamline histopathology workflows, including label-free methods that aim to reduce reliance on chemical staining. Among these, deep learning-based label-free virtual staining uses images of unstained tissue to computationally generate stained images. Despite growing interest, model development and validation remain constrained by the scarcity of paired, clinically annotated whole-slide imaging datasets. We present BRIDGE, a dataset of prostate core-needle biopsies collected at the Prof. Franciszek Łukaszczyk Oncology Centre in Bydgoszcz, Poland. The dataset comprises 869 glass slides from 169 patients, each digitized sequentially after sectioning, deparaffinization, and hematoxylin and eosin staining, yielding in total 2607 whole-slide images. The dataset supports the development and external validation of label-free virtual staining, virtual unstaining, image-registration and segmentation methods. Each slide is accompanied by a Gleason score, enabling downstream analyses of cancer classification and grading concordance. The stained images can be used independently for color normalization and cross-scanner robustness studies. All images are de-identified, distributed in their native scanner format and will be available through the BioImage Archive.}

\keywords{Stain-free histopathology, Virtual staining, Whole-slide imaging, Prostate cancer, Computational pathology, Digital pathology}

\maketitle

\section*{Background}
Histopathological diagnosis traditionally relies on tissue sectioning and histochemical staining, most commonly with hematoxylin and eosin (H\&E), to reveal tissue architecture and cellular morphology~\cite{chan2014}. However, staining adds time, reagents and laboratory processing, motivating the development of stain-free approaches that aim to simplify the conventional workflow.

Deep learning-based label-free virtual staining computationally generates images resembling conventional histochemical stains from unstained tissue images. These approaches use generative models to learn a mapping between label-free tissue images and histologically stained counterparts~\cite{latonen2024}. Because the tissue is imaged before chemical staining, the same section remains available for subsequent histological processing, special stains, molecular assays or sequencing-based analyses. Stain-free approaches may therefore reduce reagent consumption and processing time while preserving limited tissue material for complementary analyses.

Several imaging modalities have been explored for label-free virtual staining, including tissue autofluorescence microscopy~\cite{rivenson2019autofluorescence}, quantitative phase imaging~\cite{rivenson2019phasestain}, fluorescence-lifetime imaging~\cite{wang2024} or photoacoustic microscopy~\cite{yoon2024}. However, these techniques typically require specialized imaging equipment that is not routinely available in pathology laboratories, potentially limiting their integration into existing clinical workflows. In contrast, brightfield imaging can leverage conventional whole-slide scanners already widely used in digital pathology.

Label-free virtual staining from unstained brightfield images has been investigated in several tissues, including skin biopsies~\cite{hussain2025}, prostate core biopsies~\cite{rana2020}, brain, lymph nodes and liver tissue~\cite{misra2026labelfree}. Despite promising demonstrations, studies have generally relied on relatively small cohorts collected at a single institution and under a single acquisition protocol, with limited independent external validation.

Dataset size and diversity are particularly important for generative models, which must capture both the transformation between unstained and stained tissue and variation in morphology and image acquisition. Insufficient representation of these factors can lead to overfitting, unstable training and poor generalization. Scans of the same unstained and stained tissue sections are especially valuable for supervised approaches because, after registration, they provide corresponding tissue structures without the anatomical differences introduced by adjacent sections.

Robust generalization is particularly important for label-free virtual staining because it is intended as a replacement-level system, in which the generated image substitutes for the physically stained slide and is subsequently interpreted by a pathologist or analyzed by downstream AI systems for classification, grading, stratification or other clinically relevant tasks~\cite{misra2026standardization}. Hallucinations introduced during image generation may propagate to subsequent analyses and clinical decisions. Replacement-level systems thus require substantially stronger validation across institutions, scanners and tissue-processing protocols, and ultimately regulatory-grade evidence. Current label-free virtual staining studies often do not provide this level of evidence, with validation largely restricted to data from a single institution and acquisition setting. Meeting these requirements is challenging in practice because paired brightfield datasets of human unstained and stained tissue are rarely publicly available, necessitating new collections. To our knowledge, the only openly accessible human tissue dataset is E-Staining DermaRepo, comprising 87 paired unstained and H\&E-stained skin-biopsy slides from 22 patients~\cite{asaf2024}.

To address this gap, we present the BRIDGE dataset, a clinically annotated dataset of human prostate core-needle biopsies collected at the Prof. Franciszek Łukaszczyk Oncology Centre in Bydgoszcz, Poland. The dataset comprises 869 physical glass slides from 169 patients, with each section digitized sequentially at three stages of histological processing: immediately after sectioning with paraffin retained, after deparaffinization but before chemical staining, and after conventional H\&E staining. The complete collection therefore contains 2607 whole-slide images (WSIs). This design permits direct comparison of the tissue before and after paraffin removal and staining while avoiding the biological mismatch inherent to images acquired from adjacent sections.

The dataset can be used both to develop label-free virtual staining and virtual-unstaining models and as an external dataset for evaluating models trained on different patient populations, scanners or tissue-preparation protocols. The unstained WSIs can additionally support the development and evaluation of segmentation models that identify the tissue area directly from unstained brightfield images, an important preprocessing step for downstream stain-free computational pathology pipelines. The availability of Gleason scores further enables assessment of the clinical applicability of stain-free approaches, including evaluation of computational grading performance and concordance with reported pathological grades. The H\&E images can also be used independently of the unstained images for studies of prostate cancer classification, color normalization and the robustness of computational pathology models across scanners.

By releasing paired WSIs of the same prostate tissue sections at three histological processing stages together with diagnostic annotations, the BRIDGE dataset aims to facilitate the development and independent evaluation of robust stain-free computational pathology methods.

\section*{Methods}

\subsection*{Ethical considerations}
The patient sample collection was approved by the Bioethical Committee of Politechnika Bydgoska in Bydgoszcz, Poland (permit WM.0025.13.2026KB). Given the retrospective nature of the study and the use of de-identified prostate specimens, the ethics committee did not require informed consent. Access to and processing of the data at Karolinska Institutet were approved by the Swedish Ethical Review Authority (permits 2019-05220, 2025-07557-02). The samples were de-identified before transfer to Karolinska Institutet.

\subsection*{Study design}
\begin{figure}[!htb]
\centering
\includegraphics[width=\textwidth]{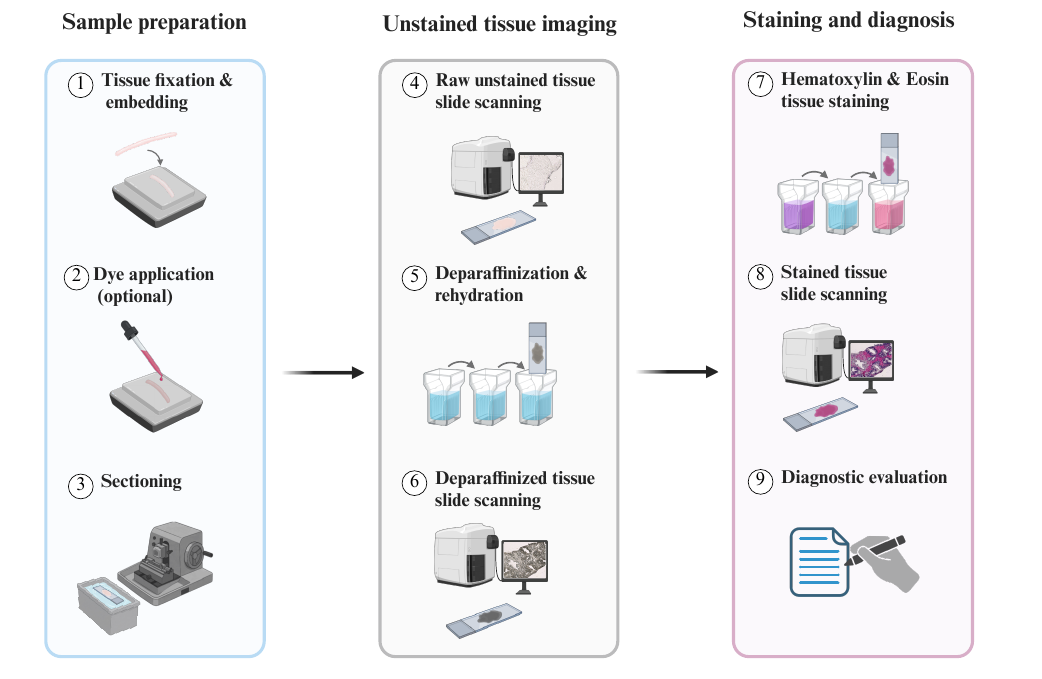}
\caption{Overview of the sample collection pipeline.}\label{fig1}
\end{figure}

This study involves retrospective sample collection of prostate core-needle biopsies, laboratory slide preparation, whole-slide scanning at three stages of tissue preparation and pathologist annotation on the stained image, as summarized in Figure~\ref{fig1}.

\subsection*{Laboratory slide processing and scanning}
\begin{figure}[!htb]
\centering
\includegraphics[width=\textwidth]{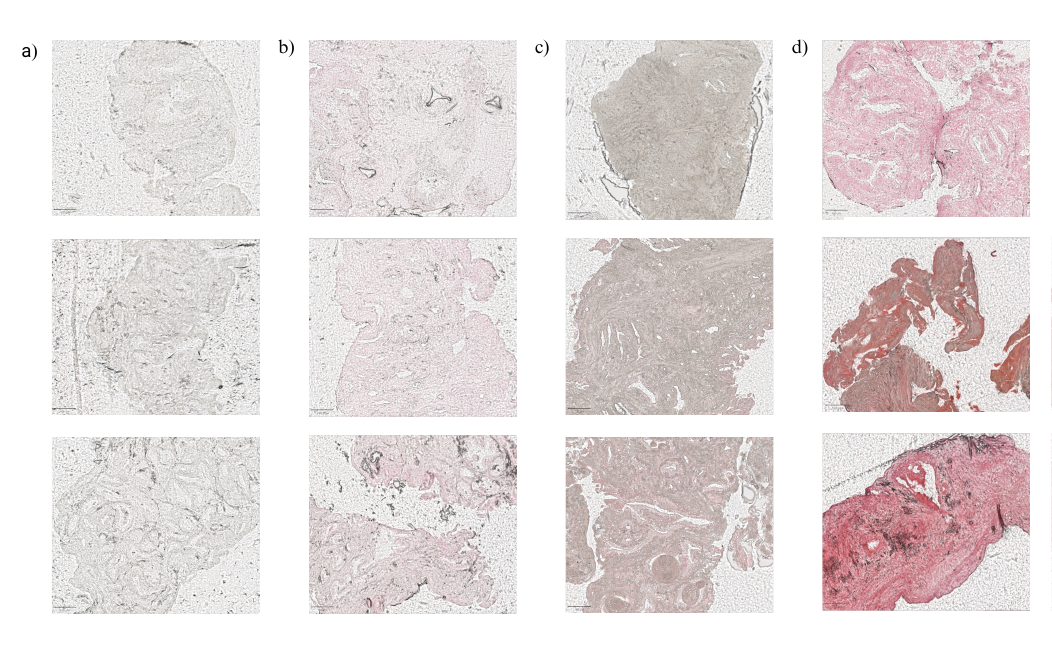}
\caption{Variation in the appearance of the raw tissue sections due to Safranin O application during paraffin embedding. a) no dye, b) faint pink tint, c) pink-brownish tint, d) intense dye color.}\label{fig2}
\end{figure}

Formalin-fixed tissue specimens were processed and embedded in paraffin. In some cases, Safranin O dye was added during paraffin embedding to improve the tissue visibility and facilitate sectioning. The amount and concentration of the dye were not standardized, which resulted in variation of the dye intensity and the raw tissue appearance (Figure~\ref{fig2}). The resulting formalin-fixed, paraffin-embedded (FFPE) tissue blocks were sectioned at a thickness of 3.5~\textmu m and mounted onto glass slides. After sectioning and initial drying, the slides were scanned prior to deparaffinization, generating WSIs. These images represented raw tissue sections prior to any chemical processing.

Following the initial scan, the slides were incubated in a drying oven at 60~\textdegree C for 30 minutes and subsequently deparaffinized in two consecutive xylene baths for 5 minutes each. Afterwards, the slides were allowed to air-dry for approximately 3 minutes to remove residual xylene and were rescanned to generate deparaffinized WSIs.

The slides were subsequently subjected to routine automated H\&E staining using a Tissue-Tek Prisma\textsuperscript{\textregistered} Plus automated slide stainer and Tissue-Tek\textsuperscript{\textregistered} H\&E Staining Kit (Sakura Finetek). The staining procedure was performed according to the validated Sakura H\&E staining protocol and comprised rehydration, hematoxylin staining, differentiation, bluing, eosin counterstaining, dehydration, clearing, and coverslipping. Following H\&E staining and coverslipping, the slides were scanned for a third time to obtain the corresponding H\&E-stained WSIs.

\begin{figure}[!htb]
\centering
\includegraphics[width=\textwidth]{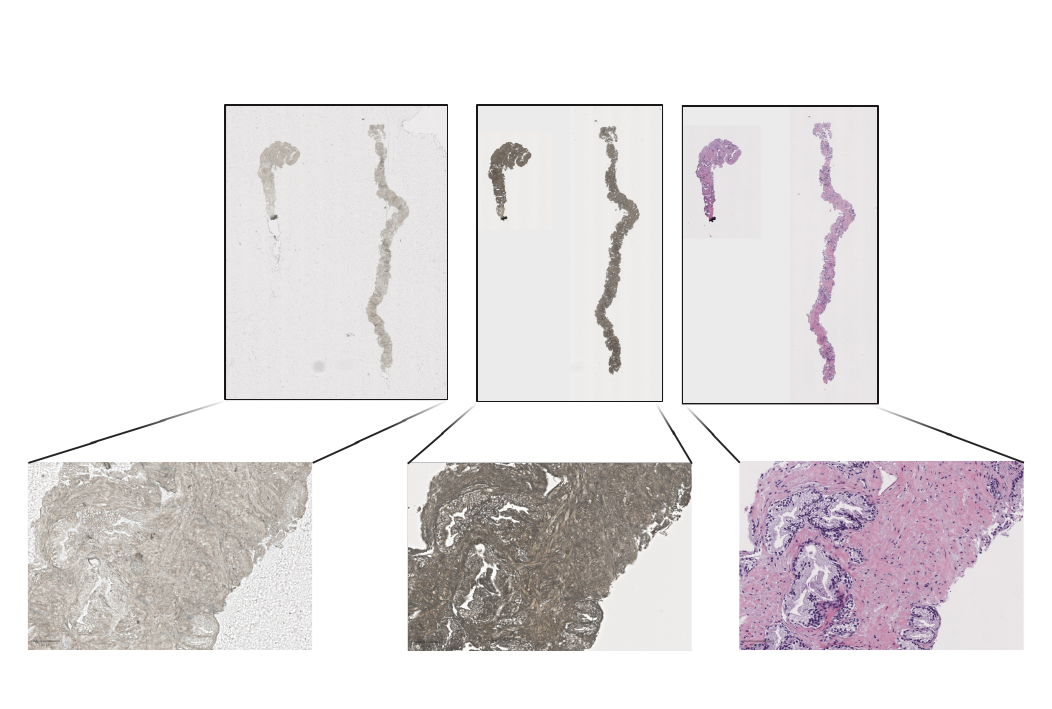}
\caption{Example of a glass slide scanned at three stages of tissue processing: raw, deparaffinized, H\&E-stained. Whole-slide thumbnails and regions at 10x magnification are shown for each stage.}\label{fig3}
\end{figure}

All slide scans at the three processing stages were generated using the VENTANA DP 600 Slide Scanner (Roche Diagnostics; serial no. DP6000236) at 40\texttimes{} magnification with a resolution of 0.25~\textmu m/pixel and stored in a TIFF-based format. An example of scans resulting from one glass slide is presented in Figure~\ref{fig3}.

The dataset includes rare modalities such as raw unstained tissue brightfield images, for which very little public data is available and thus for which there is less established intuition about image quality. Assessment of the scan quality, such as identification of the out-of-focus regions, is more challenging for these images due to low contrast and more unconventional tissue appearance compared to established H\&E-stained images. Moreover, the quality of unstained tissue imaging may be sensitive to the scanner vendor, scanner model, and acquisition settings, as these systems can be optimized for conventionally stained histology slides. Each scan was therefore assigned a quality label through visual inspection. The labels represent a subjective, descriptive assessment and are reported both to facilitate comparison of image quality across datasets and scanners and to raise awareness of potential scanner-specific limitations when imaging unstained tissue. We hope that reporting these observations will contribute to a broader discussion within the stain-free imaging community and help establish more systematic approaches to evaluating image quality for this emerging modality. Further details on the quality and dye labels are provided in the Data records section.

\subsection*{Reference standard protocol}
The digitized slides were reviewed by M.Z., a certified uropathologist from Prof. Franciszek Łukaszczyk Oncology Centre in Bydgoszcz with 5 years of diagnostic experience. Each H\&E-stained whole-slide image was assessed using Roche VENTANA uPath Digital Pathology software and assigned a slide-level Gleason score. Slides without evidence of malignancy were labeled as benign (0+0). The Gleason scores were then mapped to the corresponding ISUP Grade Groups. No additional pixel-level annotations were provided. A total of 869 H\&E-stained WSIs were evaluated, with the final labels associated with the respective de-identified slide IDs.

\begin{table}[!htbp]
\caption{Patient and slide characteristics of the BRIDGE dataset.}\label{tab1}
{\fontfamily{phv}\fontsize{9}{11}\selectfont
\renewcommand{\arraystretch}{1.9}
\setlength{\tabcolsep}{6pt}
\arrayrulecolor[gray]{0.55}
\setlength{\arrayrulewidth}{0.4pt}
\begin{tabularx}{\textwidth}{|>{\hsize=0.85\hsize}X|>{\hsize=0.75\hsize}X|>{\hsize=1.55\hsize}X|>{\hsize=0.85\hsize}X|}
\hline
\multicolumn{2}{|l|}{\textbf{Number of patients}} & \multicolumn{2}{l|}{\textbf{Number of slides}} \\
\hline
\multicolumn{2}{|l|}{\textbf{n = 169}} & \multicolumn{2}{l|}{\textbf{n = 869}} \\
\hline
\multicolumn{2}{|l|}{\textbf{Age, years}} & \multicolumn{2}{l|}{\textbf{ISUP grade}} \\
\hline
$\leq$ 49 yrs & 2  & Benign (0+0) & 502 \\
\hline
50--54 yrs & 6  & ISUP 1 (3 + 3) & 102 \\
\hline
55--59 yrs & 18 & ISUP 2 (3 + 4) & 78 \\
\hline
60--64 yrs & 32 & ISUP 3 (4 + 3) & 84 \\
\hline
65--69 yrs & 37 & ISUP 4 (4 + 4, 3 + 5, 5 + 3) & 58 \\
\hline
$\geq$ 70 yrs & 74 & ISUP 5 (4 + 5, 5 + 4, 5 + 5) & 35 \\
\hline
 & & Neuroendocrine cancer & 10 \\
\hline
\end{tabularx}
\arrayrulecolor{black}}
\end{table}

\section*{Data records}
The deposited data are organized within a single root directory, ``BRIDGE'', containing three subfolders corresponding to the consecutive tissue-processing stages: component\_A\_raw, component\_B\_deparaffinized, and component\_C\_HEstained. These folders contain WSIs acquired before deparaffinization, after deparaffinization but before staining, and after H\&E staining, respectively. In total, the dataset contains 869 glass slides from 169 patients.

A common non-identifying slide ID is retained across the three processing stages, allowing the corresponding images to be matched directly. File names follow the format \textless{}slide\_id\textgreater{}\_\textless{}stage\textgreater{}.\textless{}ext\textgreater{}, where the stage suffix indicates the tissue-processing state: \_1 for raw, non-deparaffinized tissue, \_2 for deparaffinized unstained tissue, and \_3 for H\&E-stained tissue.

WSI-level metadata is provided in BRIDGE\_metadata.csv. The table includes patient identifier (Patient\_ID), patient age at biopsy (Age), glass slide identifier (Slide\_ID), WSI identifier (WSI\_ID representing slide identifier followed by the processing stage suffix), processing stage identification (Processing\_stage), slide-level Gleason score (GS), corresponding ISUP Grade Group (ISUP) and binary tumor label (Tumor). The table also reports the acquisition quality of each scan (Scan\_quality). For raw tissue scans, the dye intensity (Dye\_intensity) is additionally provided.

Dye intensity was recorded on a four-level scale: 0 (no dye applied), 1 (faint pink tint), 2 (pink-brownish tint) and 3 (intense dye color) as illustrated in Figure~\ref{fig2}. Scan quality was labeled as high (high level of detail), medium (minor out-of-focus areas or artifacts), low (unclear quality, seemingly largely out-of-focus) or unacceptable (fully out-of-focus).

\section*{Data usage}
The WSIs can be viewed using digital pathology software. Typical preprocessing steps for computational analysis may include tissue segmentation, image tiling and, where appropriate, stain normalization.

\backmatter

\section*{Data availability}
The complete collection will be made publicly available through the BioImage Archive upon publication, and the accession number will be provided in the final version of the manuscript. A summary of the dataset and its documentation are available at \url{https://github.com/JAMielcarz/BRIDGE-dataset}. The data will be distributed under the Creative Commons Attribution 4.0 International (CC BY 4.0) license. All data are de-identified and contain no direct personal identifiers. Images and metadata can be linked to each other through the anonymized patient and slide identifiers provided with the dataset.

\section*{Code availability}
No custom code was used to create or process the dataset. Slide scanning was performed using the scanner software provided by the manufacturer.

\section*{Author contributions}
J.A.M. conceptualized the study, curated and processed the data, and drafted the manuscript.\\
J.D. oversaw the data collection including sample preparation and scanning.\\
M.M. contributed to laboratory processing of the samples.\\
M.Z. provided slide-level annotations.\\
A.A. and M.E. managed the ethical approvals and data access at Karolinska Institutet.\\
Ł.S. and A.C. supervised the work.\\
All authors read and approved the manuscript.

\section*{Competing interests}
The authors declare no competing non-financial interests. The following competing financial interests are declared: A.C. is a shareholder of Clinsight AB and has received honoraria from Astellas Pharma AB. M.E. is a shareholder of Clinsight AB and A3P Biomedical AB, and holds patents in A3P Biomedical AB.

\section*{Funding}
This research received no specific grant from any funding agency in the public, commercial or not-for-profit sectors.

\bibliography{bibliography}

\end{document}